\documentclass[lettersize,journal]{IEEEtran}
\usepackage{amsmath,amsfonts}
\usepackage{algorithmic}
\usepackage{algorithm}
\usepackage{array}
\usepackage[caption=false,font=normalsize,labelfont=sf,textfont=sf]{subfig}
\usepackage{textcomp}
\usepackage{stfloats}
\usepackage{url}
\usepackage{verbatim}
\usepackage{graphicx}
\usepackage{cite}

\usepackage{booktabs}
\usepackage{tabularx}
\usepackage{makecell}
\usepackage{multirow}
\usepackage{colortbl}
\usepackage{xcolor}
\usepackage{tablefootnote}
\usepackage{enumerate}
\usepackage{enumitem}
\definecolor{danred}{rgb}{0.9098,0.9098,0.9098}

\begin{document}

% \title{AnyControl: A Universal Framework for Controllable Video Generation with Any Conditional Control}
% \title{A Universal Framework for Controllable Video Generation and Video Interpolation}
% \title{EasyControl: Adding controls  \\ for Video Generation and Interpolation}
% \tiny
\title{\fontsize{18pt}{20pt}\selectfont EasyControl: Transfer ControlNet to Video Diffusion for \\ Controllable Video Generation and Interpolation}
% \title{\fontsize{18pt}{20pt}\selectfont EasyControl: A Frame Number-agnostic Adapter for \\ Controllable Video Generation and Interpolation}
% \title{EasyControl: Transfer ControlNet to Video Diffusion for Controllable Generation and Interpolation}
% \normal

% controllable video generation and interpolation framework.

% \author{Cong~Wang$^{1}$\quad Jiaxi~Gu$^{2}$\quad Panwen~Hu$^{3}$\quad Songcen~Xu$^{2}$\quad Hang~Xu$^{2}$\quad Xiaodan~Liang $^{1}$\footnotemark[1] \\
% $^{1}$ Shenzhen Campus of Sun Yat-Sen University, China \\
% $^{2}$Huawei Noah’s Ark Lab, China  \quad $^{3}$
% Mohamed bin Zayed University of Artificial Intelligence\\
% {\tt\small $^{1}$\{wangc39@mail2., liangxd9@mail.\}sysu.edu.cn\quad} \\ {\tt\small $^{2}$\{imjiaxi, chromexbjxh\}@gmail.com} {\tt\small $^{3}$\{panwen.hu\}@mbzuai.ac.ae}
% }

\author{Cong~Wang, Jiaxi~Gu, Panwen~Hu, Haoyu~Zhao, Yuanfan~Guo, Jianhua~Han, Hang~Xu, Xiaodan~Liang\IEEEauthorrefmark{1}
\IEEEcompsocitemizethanks{
        \emph{
        \IEEEauthorrefmark{1} Xiaodan~Liang is the corresponding author.
        }
        \protect\\
        \IEEEcompsocthanksitem  Cong Wang is with the School of Intelligent Systems Engineering, Shenzhen Campus of Sun Yat-sen University, Shenzhen 518107, China (e-mail: wangc39@mail2.sysu.edu.cn).
        \IEEEcompsocthanksitem  Panwen Hu is with the School of Science and Engineering at the Chinese University of Hong Kong, Shenzhen (e-mail: panwenhu@link.cuhk.edu.cn).
        \IEEEcompsocthanksitem Haoyu Zhao is with the School of Computer Science and Technology, Fudan University, Shanghai, China (e-mail: hyzhao22@m.fudan.edu.cn)
        \IEEEcompsocthanksitem Jiaxi~Gu, Yuanfan~Guo, Jianhua~Han, and Hang~Xu are with Huawei Noah'ark Lab, Shanghai 201206, China (e-mail: imjiaxi@gmail.com; hanjianhua4@huawei.com; chromexbjxh@gmail.com).
        \IEEEcompsocthanksitem Xiaodan Liang is with the School of Intelligent Systems Engineering, Shenzhen Campus of Sun Yat-sen University, Shenzhen 518107, China,  DarkMatter AI Research, Guangzhou 511458, China and Pengcheng Lab, Shenzhen 518000 (e-mail: liangxd9@mail.sysu.edu.cn).
        }       
}


\markboth{Journal of \LaTeX\ Class Files,~Vol.~14, No.~8, August~2024}%
{Shell \MakeLowercase{\textit{et al.}}: A Sample Article Using IEEEtran.cls for IEEE Journals}

% % column for its text to clear the IEEEpubid mark.

\maketitle

\begin{abstract}
Following the advancements in text-guided image generation technology exemplified by Stable Diffusion, video generation is gaining increased attention in the academic community. However, relying solely on text guidance for video generation has serious limitations, as videos contain much richer content than images, especially in terms of motion. This information can hardly be adequately described with plain text. Fortunately, in computer vision, various visual representations can serve as additional control signals to guide generation. With the help of these signals, video generation can be controlled in finer detail, allowing for greater flexibility for different applications. Integrating various controls, however, is nontrivial. In this paper, we propose a universal framework called \textit{EasyControl}. By propagating and injecting condition features through condition adapters, our method enables users to control video generation with a single condition map. With our framework, various conditions including raw pixels, depth, HED, etc., can be integrated into different Unet-based pre-trained video diffusion models at a low practical cost. We conduct comprehensive experiments on public datasets, and both quantitative and qualitative results indicate that our method outperforms state-of-the-art methods. EasyControl significantly improves various evaluation metrics across multiple validation datasets compared to previous works. Specifically, for the sketch-to-video generation task, EasyControl achieves an improvement of 152.0 on FVD and 19.9 on IS, respectively, in UCF101 compared with VideoComposer. For fidelity, our model demonstrates powerful image retention ability, resulting in high FVD and IS in UCF101 and MSR-VTT compared to other image-to-video models. 

% Further details and comprehensive results of our model will be presented in \url{https://sense39.github.io/DreamVideo/}.

\end{abstract}


\section{Introduction}
\label{sec:intro}
\IEEEPARstart{D}{ue} to its potential applications in artistic creation, entertainment, and beyond, image and video generation have garnered significant attention. Text-guided image generation has witnessed explosive growth, with numerous excellent works published, such as GLIDE ~\cite{saharia2022photorealistic}, Imagen ~\cite{ho2022imagen}, Stable Diffusion ~\cite{rombach2022high}, among others. Building on existing text-guided image synthesis technologies, text-to-video (T2V) generation~\cite{singer2022make} has also made strides. T2V generation can be viewed as a straightforward extension of text-to-image synthesis into the video domain, achieved by incorporating temporal layers into the network while keeping spatial layers frozen. However, the ability of text-guided video generation is limited in real-world applications. Describing content-rich videos with language is often challenging, rendering text unable to provide precise control over video content. One reason for this limitation is that text struggles to capture the dynamic nature of video content. Additionally, complex textual descriptions pose a significant challenge to the representational capabilities of the generation model, increasing the likelihood of generation failure. To address this, various methods have been proposed for generating videos guided by different conditions, including optical flows~\cite{ni2023conditional}, depth sequences~\cite{wang2023videocomposer}, dragging strokes~\cite{yin2023dragnuwa}, and et al.

\begin{figure}[!ht]
    \centering
    \includegraphics[width=1.0\linewidth]{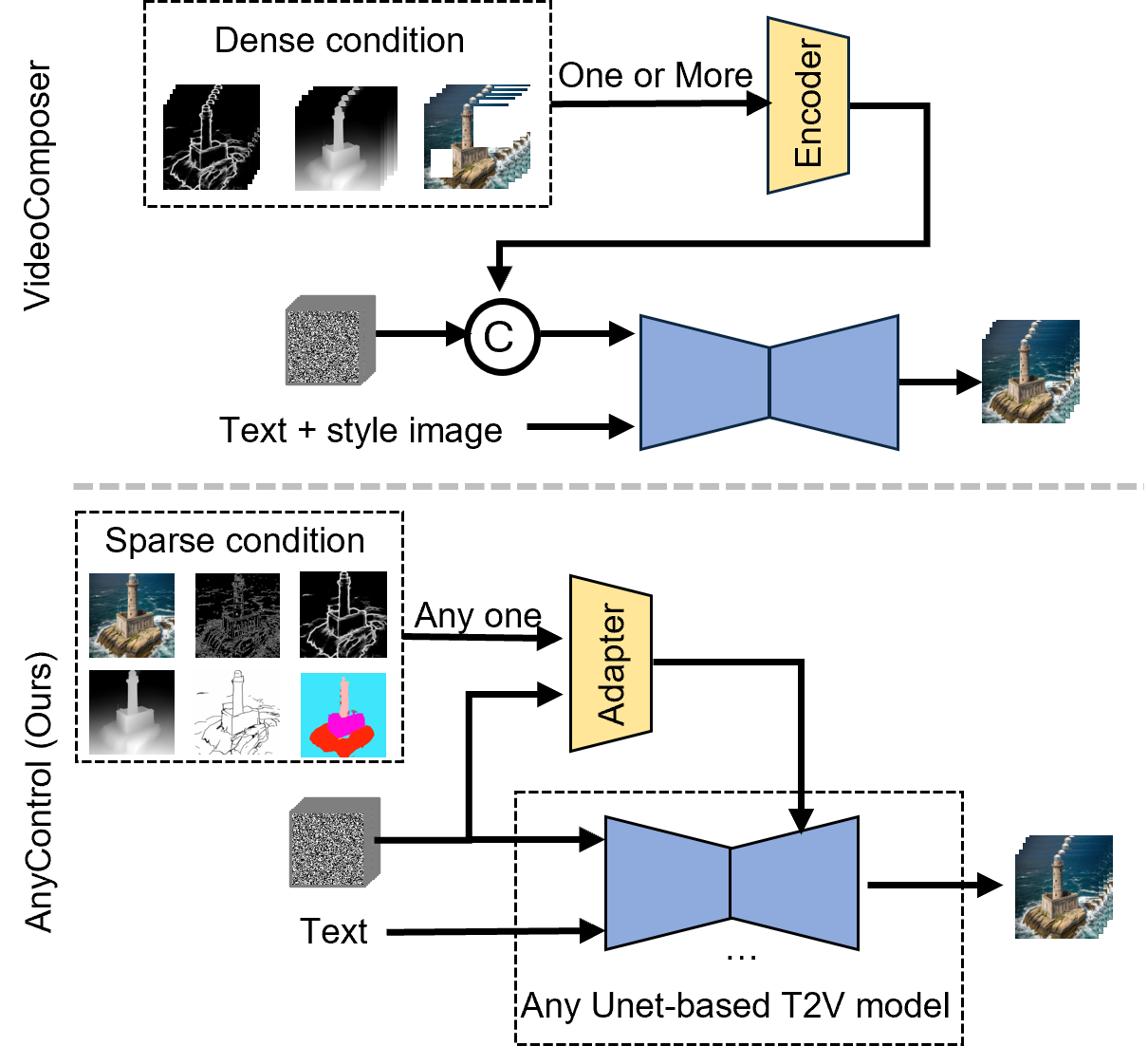}
    \caption{The architecture illustrations of a multi-condition model, VideoComposer \cite{wang2023videocomposer}, and our framework, EasyControl. Compared with VideoComposer which takes as input temporally dense conditions and injects the conditions in a concatenation manner, our EasyControl uses only a single frame of condition and injects the condition embeddings through residual summation, thereby increasing the flexibility of the framework to combine different pre-trained T2V models.   }
    \label{fig:intro}
\end{figure}

Among all the conditions, using images as a condition is a natural and intuitive way to guide video generation. Images provide substantial visual details, while text prompts can still play a role in controlling the diversity of video generation. Therefore, research on image-to-video generation with text guidance, such as~\cite{zhang2023i2vgen,guo2023i2v,ren2024consisti2v}, has attracted extensive attention. However, conditioning solely on images presents challenges due to the higher demand for fidelity and the risk of conflict between static details and smooth motion. Therefore, purely conditioning on images may lack flexibility and generalization. Hence, exploring multiple condition modalities, such as depth and sketches, is worth investigating.

Various methods have been proposed for generating videos guided by different conditions. For instance, some works introduce a sequence of condition maps~\cite{esser2023structure, zhang2023controlvideo, chen2023control}, like depth map sequences, as structural guidance to enhance temporal consistency. Moreover, a few works propose combinations of multiple conditions, such as camera and object trajectory~\cite{wang2023motionctrl}, image and optical flow~\cite{ni2023conditional}, image and trajectory~\cite{chen2023motion, yin2023dragnuwa, wu2024draganything}, and object layout and trajectory~\cite{ma2023trailblazer,chen2024motion}, to control both static visual elements and dynamic motion. By utilizing additional control signals, these methods achieve enhanced control capabilities. However, existing methods either require dense temporal condition sequences, meaning users have to provide a condition map for each frame, or they need to redesign the architecture to integrate additional condition inputs, thereby increasing practical costs. The high demand for dense condition input, the limited exploration of condition modalities, and the lack of feasibility in incorporating new control modalities result in a gap between these methods and real-world content creation processes. Hence, a question arises from these limitations: How can a method allow users to control video generation using different temporally sparse conditions, i.e., a single condition map, while lowering the training cost for each condition modality with the available well-trained text-to-video (T2V) models?

\begin{figure*}[htbp]
    \centering
  \includegraphics[width=\textwidth]{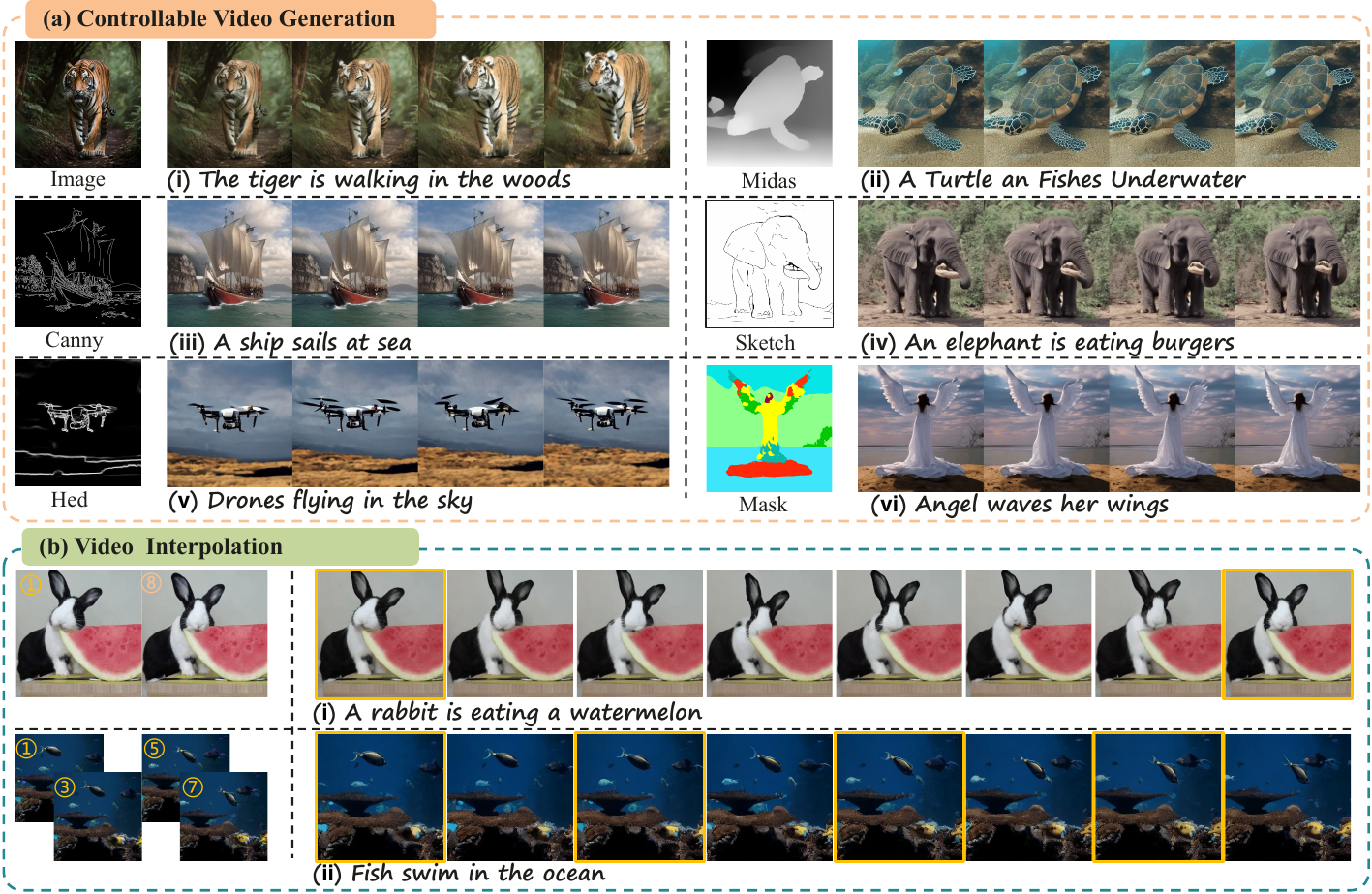}
    \caption{EasyControl is capable of generating user-defined videos by inputting any condition. Any U-Net-based text-video model can incorporate various types of input conditions through the condition adapter, such as canny, sketches, images, segment masks, and more. Users only need to provide one condition and text, and EasyControl will take care of the rest. On the left is the input condition, and on the right are the frames 1,4,5,8 of the generated videos.}
    \label{fig:showcase}
\end{figure*}

Considering the aforementioned challenges, we propose a universal framework called EasyControl for text-guided video generation with various condition modalities. In addition to the image modality, other modalities such as sketch, depth, HEDs, segmentation mask, and canny edge can be easily integrated into our framework to control video generation. Fig~\ref{fig:showcase} showcases some generation cases from various condition modalities, illustrating the different degrees of flexibility each condition provides for controllable video generation. Despite the varied conditions, text guidance retains its ability to influence the motion of the resulting video, enabling users to generate videos with varying degrees of control strength.

While existing work like VideoComposer~\cite{wang2023videocomposer} also focuses on controlling video generation with multiple conditions, it still has two main drawbacks. First, similar to previous methods~\cite{zhang2023controlvideo, chen2023control}, VideoComposer maintains temporal consistency by requiring a dense condition sequence as input, limiting its practical application feasibility. Second, its condition injection method, i.e., concatenation with latent noise, struggles to propagate control signals to all frames during generation, necessitating a redesign of the model architecture to accommodate the increased channel number resulting from concatenation injection when combined with another basic model. In contrast, our framework follows the philosophy of ControlNet~\cite{controlnet} and mainly comprises a condition adapter and an interchangeable pre-trained text-to-video (T2V) model. This design enables the incorporation of an additional condition modality by simply training the condition adapter, eliminating the need to train the entire model comprehensively. Moreover, to lower the entry barrier and allow users to control video synthesis by inputting a single condition map, we propose to propagate the condition by addition to the latent noise after extracting condition features. Subsequently, we inject the condition embedding outputted from the condition adapter through multi-layer residual summation. Experimental results also demonstrate the effectiveness of our method compared to simple concatenation.

For a comprehensive evaluation of our proposed method, we conduct extensive experiments on multiple benchmarks. In addition to assessing our method's ability to generate high-quality videos, we validate its generalization and feasibility by applying the image condition adapter on two text-to-video (T2V) models. We also investigate the results of different trained adapters for various condition modalities. On the UCF101 benchmark for image-to-video generation, we achieve a notable Fréchet Video Distance (FVD) score of 197.66, signifying a significant improvement over most existing methods (VideoCrafter1, 297.62). Moreover, we conduct comparisons with other methods from multiple perspectives, including user studies.

The contributions of our work can be summarized as follows:
\begin{itemize}
    \item We propose that EasyControl unifies both controllable video generation and video interpolation tasks in a single framework. Additionally, EasyControl is suitable for any U-Net-based video diffusion model.
    \item We design VideoInit, a noise initialization strategy that introduces the low-frequency band from input images for stable video generations.
    \item Extensive experiments demonstrate that our proposed method achieves superior quantitative results and exhibits better control capabilities compared to alternative methods.
\end{itemize}

We propose that EasyControl unifies both controllable video generation and video interpolation tasks in a single framework by residual conditions injection and reuses the low-frequency information of input images.

% We propose that EasyControl unifies both controllable video generation and video interpolation tasks in a single framework. Additionally, EasyControl is suitable for any U-Net-based video diffusion model.

% We design VideoInit, a noise initialization strategy that introduces the low-frequency band from input images for stable video generations.

% an effective noise initialization strategy for inference endows the video interpolation, which initializes Gaussian noise by introducing low-frequency input images.

% We propose a concise yet effective noise initialization strategy, referred to as VideoInit, which initializes noise by introducing low-frequency input images.

% \begin{figure*}[htbp]
%     \centering
%     \includegraphics[width=1.0\linewidth]{figures/comp.pdf}
%     \caption{The comparison in image-to-video and sketch-to-video of VideoComposer, EasyControl(VidRD) and EasyControl(ModelScope). 
%     \textit{ACtrl.}, \textit{Msp.} and \textit{Vid.Composer} denotes EasyControl, ModelScope and VideoComposer.}
%     \label{fig:comp}
% \end{figure*}
\section{Related work}
\label{sec:rel}

\subsection{Video diffusion models}

Diffusion Models (DMs)~\cite{2020DMs} have demonstrated remarkable results in image synthesis, leading to the development of various methods such as GLIDE~\cite{nichol2021glide}, Imagen~\cite{saharia2022photorealistic}, and Stable Diffusion~\cite{rombach2022high}, among others. As the field progresses, attention has shifted towards video generation, with a prevalent approach involving the integration of temporal layers into image DMs to enable temporal representation. Numerous video diffusion models have emerged, including Make-A-Video~\cite{singer2022make}, CogVideo~\cite{hong2022cogvideo}, Imagen Video~\cite{ho2022imagen}, MagicVideo~\cite{zhou2022magicvideo}, and VidRD~\cite{gu2023reuse}, among others. Regarding training data, recent works~\cite{blattmann2023align,wang2023lavie} have demonstrated that combining both image and video data can significantly enhance appearance details and mitigate catastrophic forgetting. Recognizing the complexity of video data, PVDM~\cite{yu2023video} introduces an image-like 2D latent space for efficient parameterization. Moreover, studies on the impact of initial noise priors in video DMs have been conducted. VideoFusion~\cite{luo2023videofusion} reveals that image priors from pre-trained models can be efficiently shared across all frames, facilitating the learning process. PYoCo~\cite{ge2023preserve} devises a video noise prior to achieving improved temporal consistency. Additionally, some works leverage additional DMs for tasks such as frame interpolation, prediction, and super-resolution to enhance performance. Align Your Latent~\cite{blattmann2023align} and LAVIE~\cite{wang2023lavie} are two comprehensive pipelines for generating high-quality videos. They both employ a basic video DM to generate initial video frames and incorporate additional modules for temporal interpolation and Video Super Resolution (VSR).

\subsection{Controllable video generation}

Earlier video generation works based on image diffusion models have predominantly relied on text guidance. While text prompts can lead to creative video generation, they lack precise control over appearance, layout, or motion. Consequently, recent efforts have focused on integrating other conditions or controls into video DMs. One such approach involves using an initial image to guide video generation, also known as image animation, which has garnered significant attention. Recent advancements~\cite{zhang2023i2vgen,guo2023i2v,ren2024consisti2v} in this direction suggest encoding the image condition with a separate branch~\cite{xing2023dynamicrafter} or concatenating the image latent with the noise input~\cite{chen2023seine}. However, generating high-fidelity videos with image control remains challenging due to static visual details in the image. Additionally, other low-level representations for dense spatial structure control have been introduced. Gen-1 pioneered the use of depth map sequences as structural guidance~\cite{esser2023structure}. Similarly, ControlVideo~\cite{zhang2023controlvideo} and Control-A-video~\cite{chen2023control} attempt to generate videos conditioned on sequences of dense control signals such as edge or depth maps. VideoComposer~\cite{wang2023videocomposer} devises a unified interface for multiple conditions, leveraging temporally and spatially dense control signals to achieve fine-grained controllability. However, obtaining dense guidance signals in real-world applications is challenging and not user-friendly. Recently, several works have begun leveraging object and layout trajectory information to control the dynamic motion of synthesized videos. For instance, DragNUWA~\cite{yin2023dragnuwa} encodes sparse strokes into dense flow, which is then utilized to control the motion of objects. Similarly, motionCtrl~\cite{wang2023motionctrl} encodes the trajectory coordinates of objects into a vector map, which guides the motion of the objects. Another line of research~\cite{ma2023trailblazer, wang2024boximator, chen2024motion} focuses on achieving object motion control by encoding provided object layouts, typically represented as bounding boxes, and their corresponding trajectories. 

While existing work has explored various control conditions for controllable generation, they often require strict condition inputs or necessitate redesigning model structures and training to adapt to different condition modalities. Therefore, we propose EasyControl, a unified controllable video generation framework. This framework requires minimal overhead to train different condition adapters with pre-trained T2V models, enabling users to generate high-quality videos controllably using various accessible conditions.

\section{Method}
\label{sec:method}

% The framework proposed designs an auxiliary condition embedding module, i.e., the condition adapter, whose hidden latents can be injected into any text-to-video generation model for the control purpose, thereby endowing the model with the capability to generate videos from any given condition, including but not limited to canny-to-video, sketch-to-video, and image-to-video.
% For illustrative purposes, we apply our approach to a primary Text-to-Video (T2V) model named VdRid. The controllable video generation model shown in Fig.~\ref{fig:model}, consists of two different networks: a primary Text-to-Video (T2V) model and our Condition Adapter, which the T2V model provides the basic text-to-video generation capability and the Condition Adapter fuses conditional control signals into the U-Net~\cite{U-Net}.
% Consequently, the model possesses the capability to generate tailored videos by harmoniously merging text and various control signal inputs, thereby enabling support for a diverse array of downstream tasks. 
% The forthcoming sections will introduce some preliminary knowledge about the diffusion model, elaborate on Condition Injection block design, elucidate how to integrate the control signals into U-Net, and illustrate how inference is realized under given conditions.
The proposed framework designs an auxiliary condition embedding module, called the condition adapter, whose hidden latents can be injected into any text-to-video generation model for control purposes. This endows the model with the capability to generate videos from any given condition, including but not limited to canny-to-video, sketch-to-video, and image-to-video. For illustrative purposes, we apply our approach to a primary Text-to-Video (T2V) model named VdRid. The controllable video generation model consists of two different networks: a primary Text-to-Video (T2V) model and our Condition Adapter. The T2V model provides the basic text-to-video generation capability, while the Condition Adapter fuses conditional control signals into the U-Net. Consequently, the model possesses the capability to generate tailored videos by harmoniously merging text and various control signal inputs, thereby enabling support for a diverse array of downstream tasks. The forthcoming sections will introduce some preliminary knowledge about the diffusion model, explain the meaning of various conditions and extraction methods, elaborate on the design of the Condition Injection block, and elucidate how to integrate the control signals into the U-Net.

\begin{figure*}[!ht]
    \centering
    \includegraphics[width=1.0\linewidth]{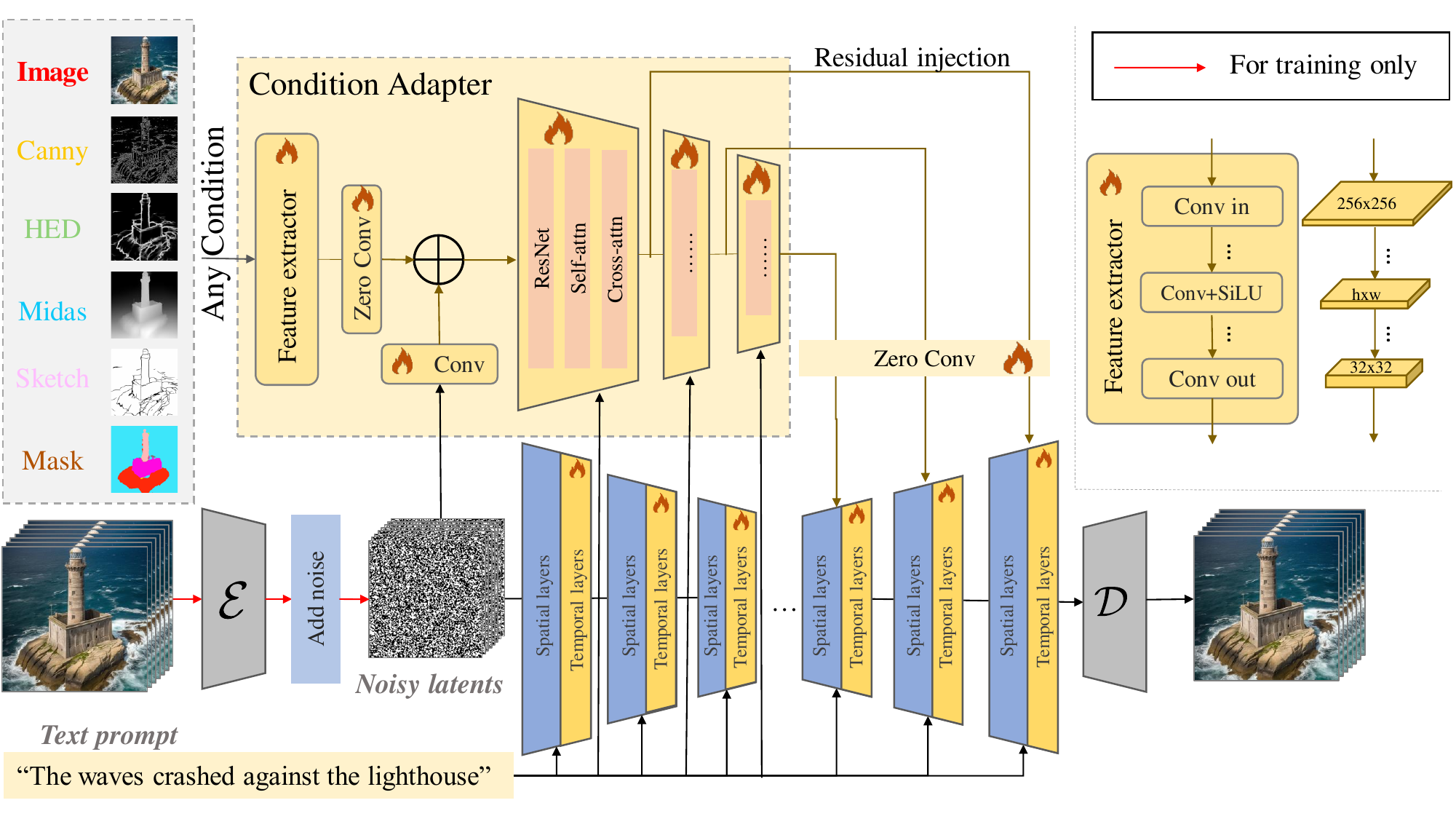}
    \caption{The EasyControl architecture encompasses the condition adapter Module, where a feature extractor block is employed to process a singular condition map, extracting pertinent condition features. These features are subsequently extended to the temporal dimension via broadcast mechanism and addition operations, incorporating noise as necessary. The integration of condition information into the generation process is achieved by augmenting the latent representations of the U-Net with multi-layer condition latents derived from the condition adapter. }
    \label{fig:model}
\end{figure*}

\subsection{Preliminary}
\textbf{Diffusion models (DMs)}~\cite{2020DMs} is a probabilistic generative model that learns the underlying data distribution through two steps: diffusion and denoising. Specifically, during the diffusion process, given an input data $\mathbf{z}$, the model gradually adds random noise $ \mathbf{z}_{t} = \alpha_{t}\mathbf{z} + \sigma_{t}\epsilon $, where $ \epsilon \in \mathcal{N}(\mathbf{0}, \mathbf{I}) $. The magnitudes of noise addition are controlled by $ \alpha_{t} $ and $ \sigma_{t} $ as the denoising steps $t$ progress. In the following denoising stage, the model takes the diffused sample $ \mathbf{z}_{t} $ as input and minimizes the mean squared error loss to learn a denoising function $ \epsilon_{\theta} $ as follows: 
\begin{equation}
    E_{\mathbf{z},\epsilon,t}=\left[ \lVert \epsilon_{\theta}(\mathbf{z}_{t}, t) - \epsilon \lVert \right]
\end{equation}

\noindent \textbf{Latent Diffusion Models (LDMs)}~\cite{LDMs} utilize the architecture of Variational Autoencoders (VAEs). Unlike Diffusion Models (DMs), LDMs can compress the input data into a latent variable $ \epsilon(\mathbf{z}) $ by encoder $ \epsilon $, and then perform denoising truncation by decoding $D(\mathbf{z}_{0})$ by the decoder $D$. LDMs significantly reduce the training and inference time as the diffusion and denoising are performed in the latent space rather than the data space. The objective of LDMs can be formulated as follows:
\begin{equation}
    E_{\mathbf{z},\epsilon,t}=\left[ \lVert \epsilon_{\theta}(\mathcal{E}(\mathbf{z}_{t}), t) - \epsilon \lVert \right]
\end{equation}

\noindent \textbf{Video Latent Diffusion Models (VLDMs)} build upon LDMs by incorporating a temporal module to capture the temporal continuity in video data. VLDMs typically add a temporal attention module to the U-Net architecture, enabling attention in the temporal dimension. Additionally, the 2d convolutions are modified to 3d convolutions to accommodate video data. Given the impressive capabilities of VLDMs in video generation, the T2V model used in our framework follows the principles of VLDMs, which encodes video into latent variables $ \mathcal{E}(\mathbf{z}) $ and leverages U-Net to learn the spatio-temporal characteristics of the video data.

\subsection{Model architecture}

% Our model architecture is shown in Figure~\ref{fig:model}. Since our model is fine-tuned based on a pre-trained T2V model, the basic structure is a latent diffusion model with text guidance through cross-attention layers. For more flexibility in generation control, a condition adapter module is designed to inject various conditions into the backbone model. For any controlling condition such as raw pixels or canny edges, a universal feature extractor, implemented with multi-scale convolutions, is used for extracting features from the input condition. 
% For condition features and noisy latents, we design a latent-ware condition propagation mechanism in the condition adapter, which can endow the inject feature condition features. The noisy latents equipped with condition features is then input into the spatial-aware block, which is initiated by the encoder of the diffusion model. The output of the condition adapter is then injected into the decoder part of the diffusion model in a residual manner.

Our model architecture, depicted in Fig.~\ref{fig:model}, is rooted in a latent diffusion framework fine-tuned from a pre-trained T2V model. This foundation incorporates text guidance via cross-attention layers. To imbue our model with enhanced generation control capabilities, we introduce a Condition Adapter Module for the integration of diverse conditions into the core model. Regardless of the controlling condition, whether it be raw pixels or canny edges, a universal feature extractor, implemented with multi-scale convolutions, is employed to extract features from the input condition. Within the Condition Adapter, we devise a latent-aware condition propagation mechanism, i.e., adding the representations of noisy latents to the condition features, to facilitate the incorporation of feature conditions into noisy latents. Subsequently, these conditioned latents are input into a spatial-aware block, initiated by the encoder of the diffusion model. The resultant output from the Condition Adapter is then injected into the decoder segment of the diffusion model in a residual fashion.

\subsection{Versatile Conditions}

% To enable the flexibility of video generation, various conditions are supported in EasyControl through a unified structure. Different from previous works like VideoComposer~\cite{wang2023videocomposer} which relies on multiple conditions simultaneously. Any of the controls can work independently through a condition adapter so higher flexibility can be achieved. Specifically, these conditions are listed below.

In EasyControl, the flexibility of video generation is facilitated by supporting various conditions through a unified structure. This approach diverges from previous methodologies, such as VideoComposer~\cite{wang2023videocomposer}, which rely on multiple conditions simultaneously. Instead, each control can operate independently through a condition adapter, thus allowing for greater flexibility. The specific conditions supported by EasyControl are enumerated below.

\begin{itemize}
    \item \textbf{Raw pixels}: This represents the most fundamental form of an image, comprising a matrix of pixel values. For the purpose of implementing image-to-video generation, we opt to utilize the initial frame of the provided video as the image condition.
    \item \textbf{Canny edges~\cite{canny}}: By detecting regions with rapid changes in intensity, it can capture both prominent and subtle edges while simultaneously minimizing noise.
    \item \textbf{HED (Holistically-Nested Edge Detection)~\cite{hed}}: By detecting the edges of an image, HED aims to capture both low-level and high-level image features. We extract the HED condition of the video frames using the HED edge detection model~\cite{hed}.
    \item \textbf{Midas~\cite{midas}}: The Midas information of the image represents the image depth information and can predict the distance of the image object from the camera.
    % the distance of objects in the scene from the camera.
    % \item \textbf{Sketch}: A sketch is a simplified representation of an image that highlights the main contours and outlines. We first get the HED boundary of the video frames. Then, we employ a sketch simplification method~\cite{sketch1,sketch2} to generate the sketches for the model training.
    \item \textbf{Sketch}: A sketch serves as a simplified depiction of an image, emphasizing its primary contours and outlines. Our approach involves obtaining the HED boundaries of the video frames initially. Subsequently, we utilize a sketch simplification method, as outlined in previous works~\cite{sketch1,sketch2}, to generate the sketches required for model training.
    % \item \textbf{Segmentation mask}: A segmentation mask is an image where each pixel is assigned a label corresponding to the object or region it belongs to. In order to efficiently generate mask labels required for training, we use ~\cite{uniformer} to label video frames.
    \item \textbf{Segmentation mask}: A segmentation mask is an image wherein each pixel is assigned a label corresponding to the object or region it pertains to. To streamline the generation of mask labels necessary for training, we employ the method proposed by Uniformer~\cite{uniformer} to label the video frames efficiently.
\end{itemize}

% These controlling conditions focus on different aspects of the visual modality. As inputs of the model, they can be represented in a unified format like RGB space. Also, for injecting features of the input controlling condition, a feature extractor is implemented.

These controlling conditions address various aspects of the visual modality. When used as inputs to the model, they can be standardized into a unified format, such as the RGB color space. Furthermore, to incorporate features from the input controlling conditions, we employ a feature extractor.

\subsection{Condition Adapter}\label{image-retention}

For our condition adapter, we take inspiration from ControlNet~\cite{controlnet}. Specifically, we copy the spatial structure and initial the weights of the condition adapter from the encoder and middle block in the diffusion model U-Net. Subsequently, 
we gradually integrate condition information into the decoder block in U-Net during the denoising process. To achieve this, we keep the spatial structure in the U-Net fixed and add the zero convolution layers at the final of the condition adapter. The output of the condition adapter is injected into the decoder block gradually.
For brevity, we denote the encoder as $E$, the middle block as $M$, and
the decoder as $D$, with $e_{i}$ and $d_{i}$ denoting the output of the $i$-th block in the encoder and decoder, and $m$ denotes the output of the middle block, respectively. 
It is important to note that, due to the
adoption of skip connections in UNet, the input for the $i$-th block in the decoder is given by:
\begin{equation}
\left\{
     \begin{aligned}
     &concat(m+m^{\prime}, e_{j} + zero(e^{\prime}_j))  \quad {where \quad  i=1, i+j=13.}  \\
     &concat(d_{i-1}, e_{j} + zero(e^{\prime}_j)) \quad {where  \quad 2 \leq i \leq 12, i+j=13.} \\
     \end{aligned}
\right.
\end{equation}
where $zero$ represents the zero convolution layer whose weights increase from zero to gradually integrate control information into the main diffusion model.
In order to extract the input condition information, we design a feature extractor block consisting of multi-convolution layers, which will increase the channel and decrease the size of the input condition denoted as $H$. The noisy latents $z$ will add the extracted condition features directly and broadcast in the time dimension:
\begin{equation}
    z^{\prime}=conv(z)+zero(H(c))
\end{equation}
where $c$ denotes the input condition.

\section{Experiments}
\label{sec:exp}

\subsection{Experimental setup}

\textbf{Implementation Details}
We apply EasyControl to two open-source Text-to-Video (T2V) generation models: VidRD~\cite{gu2023reuse} and ModelScope ~\cite{ModelScope}. To integrate EasyControl, we freeze the self-attention and cross-attention in the U-Net of each T2V model and train the temporal layer and the Condition Adapter, whose weights are initialized from the spatial layers in the downsample blocks of the U-Net. Our training process involves fine-tuning the T2V models on high-quality text video paired datasets to enhance video quality ~\cite{svd}. For this purpose, we utilize the Pexels 300K dataset ~\cite{corran2022pexelvideos}, comprising 340K video-caption pairs obtained from pexels.com without watermarks. To ensure the model's generative capability with only one input condition, we introduce a 10\% probability of replacing the condition with an empty image and a 10\% probability of empty text. During training, we conduct three epochs for each condition-to-video generation model and set 1000 warmup steps. Further training details can be found in the appendix.

\noindent \textbf{Evaluation}
The evaluation datasets are UCF101~\cite{soomro2012ucf101} with prompts sourced from VidRD~\cite{gu2023reuse} and MSR-VTT~\cite{xu2016msr}. Following previous works~\cite{singer2022make,luo2023videofusion}, we employ the following evaluation metrics: 
\romannumeral1) \textbf{Fr\'{e}chet Video Distance} (FVD)~\cite{unterthiner2019fvd}, computed by a trained I3D model~\cite{carreira2017quo}, as established in the Make-A-Video study~\cite{singer2022make}.  
\romannumeral2) \textbf{Inception Score} (IS)~\cite{saito2020train}. Following previous studies~\cite{singer2022make,hong2022cogvideo,luo2023videofusion}, we utilize a trained C3D model to compute the video version of the IS. 
\romannumeral3) \textbf{CLIP Score} measures the semantic similarity between the generated video and the corresponding prompt, where the video feature is passed through the CLIP visual encoder and the text feature is passed through the CLIP text encoder.
\romannumeral4) \textbf{Average Flow}(Avg-Flow) denotes the video motion strength. We extract optional flow by~\cite{unimatch} and calculate the average flow over frames.

% \romannumeral4) \textbf{SSIM}~\cite{SSIM} assesses the similarity between two images based on luminance, contrast, and structural integrity, which ranges from 0 to 1 and higher values denote lower image distortion.
%##################################################################################################
\begin{table*}[htbp]
    \centering
    \setlength{\tabcolsep}{2.1mm}
    % \small
    \caption{Comparison with VideoComposer, I2VGen-XL, VideoCrafter1 and SVD for zero-shot text-to-video generation on UCF101 and MSR-VTT. '-' Indicates that no measurement is required as the model generates video without text input.}
    \begin{tabular}{l c cccc cccc}
    \toprule
    \multirow{2}{*}{Methods} & \multirow{2}{*}{\#Videos} & \multicolumn{4}{c}{UCF101} & \multicolumn{4}{c}{MSR-VTT}     \\
    \cmidrule(lr){3-6} \cmidrule(lr){7-10}

    &  & FVD$\downarrow$ & IS$\uparrow$ &  CLIP Score$\uparrow$ &  Avg-Flow$\uparrow$ & FVD$\downarrow$ & IS$\uparrow$ & CLIP Score$\uparrow$ & Avg-Flow$\uparrow$  \\
    \midrule
    VideoComposer~\cite{wang2023videocomposer} & 10M & 383.87  & 34.23 & 29.60 & 14.09 & 331.15  & 12.33 & 27.03 & 12.39 \\
    I2VGen-XL~\cite{zhang2023i2vgen} & 10M & 526.94  & 18.90 & - & 14.43 & 341.72 & 10.52 & - & 12.49 \\
    VideoCrafter1~\cite{chen2023videocrafter1} & 10.3M & 297.62 & 50.88 & 24.49 &  11.45 & 201.46  & 14.41 & 21.57 & 10.04   \\
    SVD~\cite{svd} & 9.8M~\tablefootnote{The 9.8M training data is filtered from the 500M data.} & 399.59  & 45.65 & -  & 7.43 & 209.74  & 13.44 & - & 6.94 \\
    \midrule
    \rowcolor{danred} Ours & 5.3M+340k~\tablefootnote{340k refers to the amount of training data, while 5.3 M pertains to the training data used for the pre-training model.} & \textbf{197.66}  & \textbf{54.39} & \textbf{30.29}  & \textbf{16.40} & \textbf{149.18}  & \textbf{15.25} & \textbf{27.33}  & \textbf{13.25} \\
    \bottomrule
    \end{tabular}

    \label{tab:t2v metric}
\end{table*}%
%##################################################################################################

%##################################################################################################
\begin{table}[htbp]
    \centering
    \setlength{\tabcolsep}{1.3mm}
    % \small
    \caption{Comparison with VideoComposer, I2VGen-XL, VideoCrafter1 and SVD for zero-shot text-to-video generation on UCF101 and MSR-VTT. '-' Indicates that no measurement is required as the model generates video without text input.}
    \begin{tabular}{l c ccc ccc}
    \toprule
    \multirow{2}{*}{Methods}  & \multicolumn{3}{c}{Davis-7} & \multicolumn{3}{c}{UCF101-7}     \\
    % \cmidrule(lr){3-6} \cmidrule(lr){7-10}

    &  PSNR$\uparrow$ & SSIM$\uparrow$ &  LPIPS$\downarrow$ &  PSNR$\uparrow$ & SSIM$\uparrow$ &  LPIPS$\downarrow$ \\
    \midrule
    AMT~\cite{wang2023videocomposer}  & \textbf{21.09} & 0.5443  & 0.254  & xxx & xxx & xxx  \\
    RIFE~\cite{zhang2023i2vgen}  & 20.48 & 0.5112 & 0.258  & xxx & xxx & xxx  \\
    FILM~\cite{chen2023videocrafter1} & 20.71 & 0.5282 & 0.2707  & xxx & xxx & xxx  \\
    LDMVFI~\cite{svd}  & 19.98 &  0.4794 & 0.2764  & xxx & xxx & xxx  \\
    VIDIM~\cite{svd}  & 19.62 & 0.4709 & 0.2578  & xxx & xxx & xxx  \\
    \midrule
    \rowcolor{danred} Ours & 19.97 & \textbf{0.5486}  & \textbf{0.2506}  & xxx & xxx & xxx  \\
    \bottomrule
    \end{tabular}

    \label{tab:t2v metric}
\end{table}%
%##################################################################################################

%##################################################################################################
\begin{table}[htbp]
    \centering
    \setlength{\tabcolsep}{4mm}
    % \small
    \caption{Davis-7}
    \begin{tabular}{l ccc}
    \toprule
    % \multirow{2}{*}{Methods}  & \multicolumn{3}{c}{Davis-7} & \multicolumn{3}{c}{UCF101-7}     \\
    % % \cmidrule(lr){3-6} \cmidrule(lr){7-10}

   Methods &  PSNR$\uparrow$ & SSIM$\uparrow$ &  LPIPS$\downarrow$  \\
    \midrule
    AMT~\cite{wang2023videocomposer}  & \textbf{21.09} & 0.5443  & 0.254   \\
    RIFE~\cite{zhang2023i2vgen}  & 20.48 & 0.5112 & 0.258  \\
    FILM~\cite{chen2023videocrafter1} & 20.71 & 0.5282 & 0.2707 \\
    LDMVFI~\cite{svd}  & 19.98 &  0.4794 & 0.2764  \\
    VIDIM~\cite{svd}  & 19.62 & 0.4709 & 0.2578 \\
    \midrule
    \rowcolor{danred} Ours & 19.97 & \textbf{0.5486}  & \textbf{0.2506}  \\
    \bottomrule
    \end{tabular}

    \label{tab:t2v metric}
\end{table}%
%##################################################################################################

%% 大图 condtion2vidoe
%% 小的 image2video sketch2video on modelscope sdvideos videocomposer
% The frame sample rate of VideoComposer and EasyControl(ModelScope) is 2, and that of EasyControl(VidRD) is 1. 
\begin{figure*}[htbp]
    \centering
    \includegraphics[width=1.0\linewidth]{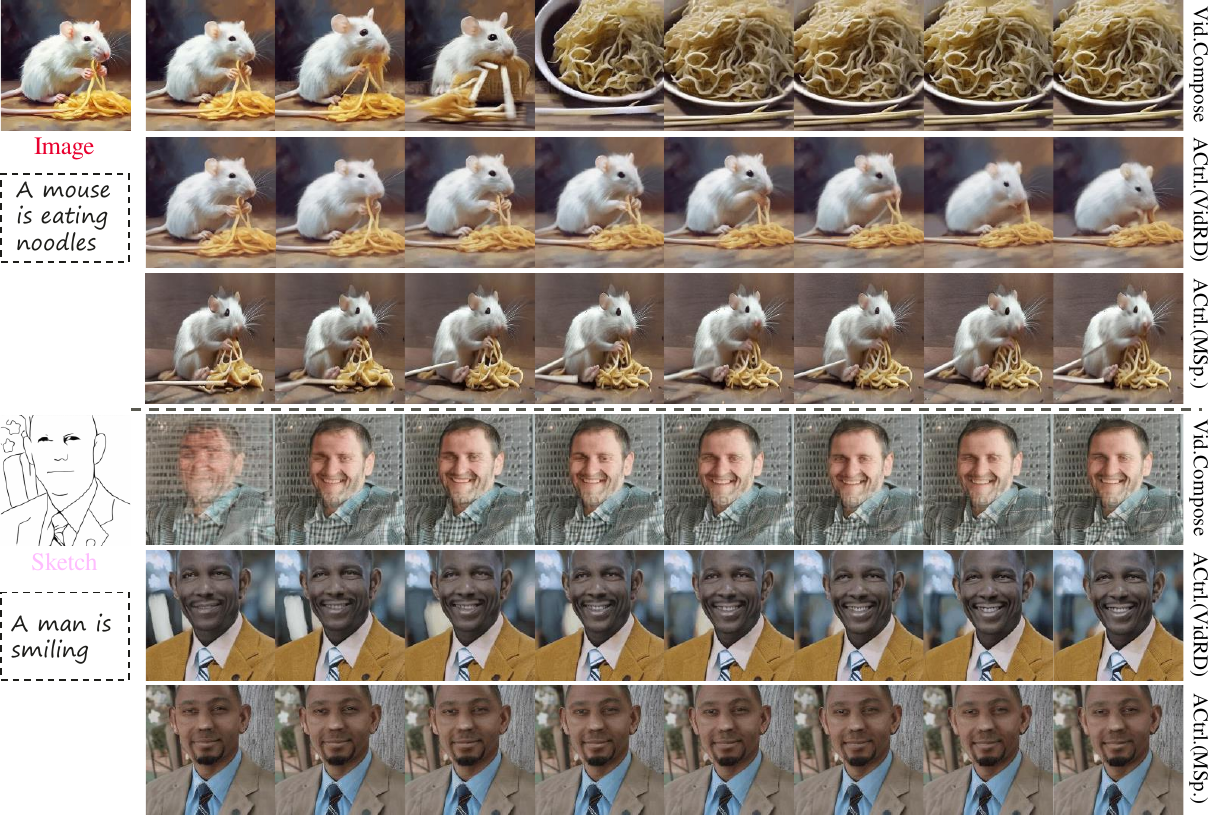}
    \caption{The comparison in image-to-video and sketch-to-video of VideoComposer, EasyControl(VidRD) and EasyControl(ModelScope). 
    \textit{ACtrl.}, \textit{Msp.} and \textit{Vid.Composer} denotes EasyControl, ModelScope and VideoComposer.}
    \label{fig:comp}
\end{figure*}
% The first part is the image-to-video 

\subsection{Qualitative evaluation}\label{qual eval}
% We conduct a qualitative comparison of our approach against VideoComposer~\cite{wang2023videocomposer} on the image-to-video and sketch-to-video tasks. We apply our method to T2V model VidRD~\cite{gu2023reuse} and ModelScope~\cite{ModelScope} named EasyControl(ModelScope) separately. 

We performed a qualitative comparison of our approach against VideoComposer~\cite{wang2023videocomposer} on both the image-to-video and sketch-to-video tasks. Specifically, we applied our method to the T2V models VidRD~\cite{gu2023reuse} and ModelScope~\cite{ModelScope}, denoted as EasyControl(VidRD) and EasyControl(ModelScope) respectively.

% \noindent \textbf{Image-to-video task}
% From Fig~\ref{fig:comp}, we find the videos generated by VideoComposer have a sudden change, where the frames of the first part match the given text prompt and image but the content of video frames becomes "A bowl of noodles". In contrast, the video generated by EasyControl(VidRD) has high fidelity of the first frame and high quality. EasyControl(VidRD) and EasyControl(ModelScope) generate videos from images both have strong continuity and match the action given in the text prompt.
% We notice the generated video quality of EasyControl(ModelScope) worse than EasyControl(VidRD), which we believe is due to the problem of the amount of training data.
% We observe that the video quality generated by EasyControl (ModelScope) is inferior to EasyControl (VidRD), which may be attributed to the limitation of the training data number of the former.
\noindent \textbf{Image-to-video task}
From Fig.~\ref{fig:comp}, we observed that the videos generated by VideoComposer exhibit a sudden change, wherein the frames of the initial segment align with the provided text prompt and image, but subsequently transition to unrelated content, such as "A bowl of noodles". In contrast, the videos generated by EasyControl(VidRD) demonstrate a high fidelity to the first frame and overall exhibit superior quality. Both EasyControl(VidRD) and EasyControl(ModelScope) consistently maintain strong continuity and align with the action described in the text prompt when generating videos from images. However, it's worth noting that the generated video quality of EasyControl(ModelScope) appears to be slightly inferior to EasyControl(VidRD), which we attribute to the differences in training hyperparameter and training data scales of different I2V basic models. 
Specifically, this may be due to the fact that the amount of training data provided for EasyControl(ModelScope) is insufficient for ModelScope, but may be sufficient for VidRD. As far as we know, the two models differ greatly in terms of the type and size of training data. Alternatively, it could be due to the influence of hyperparameters such as learning rate.

% \noindent \textbf{Sketch-to-video task}
% As depicted in Fig~\ref{fig:comp}, our EasyControl (ModelScope) and EasyControl (VidRD) generate higher-quality videos and have a high match degree of the given sketch condition. 
% Furthermore, the videos generated by EasyControl exhibit superior continuity compared to VideoComposer, which displays noticeable flickering and lower quality. Additionally, the outline of the man in the video of VideoComposer does not align with the sketch and overlooks the man's necktie shape.
\noindent \textbf{Sketch-to-video task}
As illustrated in Fig.~\ref{fig:comp}, both EasyControl (ModelScope) and EasyControl (VidRD) produce videos of higher quality and demonstrate a high level of alignment with the provided sketch condition. Moreover, the videos generated by EasyControl exhibit superior continuity compared to those generated by VideoComposer, which manifest noticeable flickering and lower quality. Notably, the outline of the man depicted in VideoComposer's video fails to accurately conform to the sketch, particularly overlooking the shape of the man's necktie.
We believe that this may be due to the weakness of the condition extractor network and condition feature inject block in VideoComposer, which is unable to extract texture information from image details and integrate the condition with noisy latents well.

\subsection{Quantitative evaluation}\label{Quantitative evaluation}
\textbf{Comparison on image-to-video generation}
% We quantitatively compare the image-to-video generation model of EasyControl with the recent I2V generation model, especially VideoComposer~\cite{wang2023videocomposer}, I2VGen-XL~\cite{zhang2023i2vgen} and VideoCrafter1~\cite{chen2023videocrafter1} and SVD~\cite{svd}.
% The I2V model primarily achieves image-to-video generation through two main approaches. One involves incorporating image information into the U-Net input, but this way is easy to lose the rich visual information in the input image. 
% I2VGen-XL, VideoCrafter1, and SVD utilize this method to preserve image visual details, which leads to weaker results in metrics such as FVD and IS compared to our approach.
% VideoComposer and our method inject image information into U-Net through the network. In this way, the image information is gradually injected into U-Net, so that the model can learn more abundant image details. 
% As shown in Tab~\ref{tab:t2v metric}, our model outperforms VideoComposer across all metrics. We attribute this superiority to the limitations of VideoComposer in representing comprehensive image information. Moreover, the simple incorporation of multiple conditions in VideoComposer may lead to confusion in handling certain scenarios, while we adapt a latent-aware condition propagation mechanism.
We conducted a quantitative comparison between the image-to-video generation model of EasyControl and recent I2V generation models, notably VideoComposer~\cite{wang2023videocomposer}, I2VGen-XL~\cite{zhang2023i2vgen}, VideoCrafter1~\cite{chen2023videocrafter1}, and SVD~\cite{svd}. The I2V models primarily achieve image-to-video generation through two main approaches. One involves directly incorporating image information into the input of the U-Net. However, this approach risks losing rich visual information present in the input image. I2VGen-XL, VideoCrafter1, SVD, and VideoComposer adopt this method to preserve image visual details, resulting in weaker performance in metrics such as FVD and IS compared to our approach. Our method injects image information into the U-Net throughout the network architecture. This gradual injection of image information enables the model to learn more comprehensive image details.

As shown in Tab.~\ref{tab:t2v metric}, our model outperforms VideoComposer across all metrics. We attribute this superiority to the limitations of VideoComposer in representing comprehensive image information. Additionally, the simple incorporation of multiple conditions in VideoComposer may lead to confusion in handling certain scenarios, whereas we employ a latent-aware condition propagation mechanism, enhancing the model's performance.

\noindent \textbf{Comparison on sketch-to-video generation}
% At the same time, we evaluate the ability to generate video from the sketch condition of our method. We measure the FVD and IS metrics on UCF101 and MSR-VTT compared to VideoComposer as shown in Tab~\ref{tab:s2v metrics}. We find the quality and continuity of generated videos by our method outperformed VideoComposer largely, which also can be discovered in Fig~\ref{fig:comp}. We contribute this performance enhancement to the design of the condition adapter compared to the STC-encoder in VideoComposer.
Simultaneously, we assessed the capability to generate videos from sketch conditions using our method. We evaluated the FVD and IS metrics on the UCF101 and MSR-VTT datasets compared to VideoComposer, as illustrated in Tab.~\ref{tab:s2v metrics}. We observed that the quality and continuity of the generated videos by our method significantly surpassed those generated by VideoComposer, as evident in Fig.~\ref{fig:comp}. We attribute this performance enhancement to the design of the condition adapter in our method, which provides a more effective mechanism compared to the STC-encoder utilized in VideoComposer.

%##################################################################################################
\begin{table}[htbp]
    \centering
    \setlength{\tabcolsep}{3mm}
    % \small
    \caption{Comparison of our model with VideoComposer in sketch-to-video generation on UCF101 and MSR-VTT.}
    \begin{tabular}{l cc cc}
    \toprule
    \multirow{2}{*}{Methods} & \multicolumn{2}{c}{UCF101} & \multicolumn{2}{c}{MSR-VTT}     \\
    \cmidrule(lr){2-3} \cmidrule(lr){4-5}
     & FVD$\downarrow$ & IS$\uparrow$  & FVD$\downarrow$ & IS$\uparrow$    \\
    \midrule
    VideoComposer & 538.76  & 26.43  & 558.50  & 14.02   \\
    Ours & \textbf{386.76}  & \textbf{46.33} & \textbf{306.29}  & \textbf{16.61}      \\
    \bottomrule
    \end{tabular}

    \label{tab:s2v metrics}
    \vspace{-6pt}
\end{table}%
%##################################################################################################

% \begin{figure*}[!ht]
%     \centering
%     \includegraphics[width=1.0\linewidth]{figures/comp.pdf}
%     \caption{The comparison in image-to-video and sketch-to-video of VideoComposer, EasyControl(VidRD) and EasyControl(ModelScope). The frame sample rate of VideoComposer and EasyControl(ModelScope) is 2, and that of EasyControl(VidRD) is 1. \textit{ACtrl.},\textit{Msp.} and \textit{Vid.Composer} denotes EasyControl, ModelScope and VideoComposer.}
%     \label{fig:comp}
% \end{figure*}

\subsection{Ablation study}

\begin{figure}[ht]
    \centering
    \includegraphics[width=0.9\linewidth]{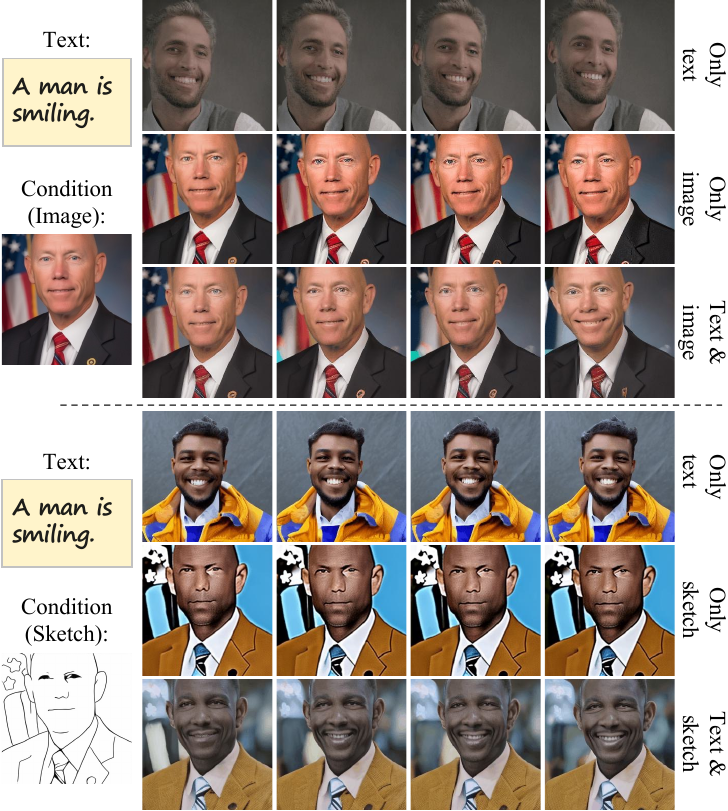}
    \vspace{-10pt}
    \caption{Image and sketch are employed for ablation studies, corresponding to the upper and lower parts of this figure, respectively. For each part, three sets of experiments are conducted using text only, condition only, and text with condition. Given frames 1,4,5,8 of the generated videos.}
    \label{fig:ablation}
    \vspace{-12pt}
\end{figure}
% \vspace{-3mm}

Given the support for various conditions within our framework, conducting ablation studies on controllable video generation becomes essential. We conducted a series of ablation studies utilizing image and sketch conditions as examples. For these studies, we utilized VidRD as the foundational T2V model in both tasks. Specifically, in line with the training set, when only the input text is provided, we set the condition as a black image. Similarly, when lacking text inputs, we set the text as empty. This experimental setup allows us to independently control the conditions and text guidance, enabling us to discern their individual effects on video generation.

% Figure~\ref{fig:ablation} shows the results of the two experiments using image and sketch conditions, corresponding to the upper and lower parts of the figure. Without the given condition, video generation with text only results in diverse generated video content. The possible reason is that the amount of information in the text is insufficient for controlling the generation of visual modality data, especially videos. On the contrary, without text guidance, the results of only using an image or sketch as a condition show high fidelity but low diversity in actions. The generated frames look very similar no matter image or sketch is used. The reason is that the visual condition is only about a static scene or character, so it can barely indicate a dynamic motion. The best generation results can be achieved with both text guidance and visual conditions. The condition of the image or sketch provides a global static appearance while it lacks the flexibility of motion control. The incorporation of text signals can endow the generated results with more possibilities, especially in terms of actions.
Fig.~\ref{fig:ablation} illustrates the outcomes of two experiments conducted with image and sketch conditions, presented in the upper and lower sections of the figure respectively. In the absence of a provided condition, video synthesis solely guided by text yields a varied array of generated video content. This variability is attributed to the limited information contained within the text for effectively controlling the generation of visual data, particularly in the context of videos. Conversely, when excluding text guidance, utilizing only an image or sketch as a condition results in high-fidelity but low-diversity outcomes in terms of actions. The generated frames exhibit remarkable similarity regardless of whether an image or sketch is employed. This uniformity arises from the static nature of the visual condition, which inadequately captures dynamic motion. Optimal generation results are achieved when both text guidance and visual conditions are utilized. While the image or sketch condition offers a static global appearance, it lacks the capacity for motion control. The integration of text signals introduces greater flexibility to the generated outcomes, particularly in terms of actions.

\subsection{User Study}
\textbf{Base model setting}
% We conduct a user study to assess our EasyControl on image-to-video and sketch-to-video tasks. We use VidRD as the basic T2V model, which can generate 8-frame videos once given conditions or text prompts at $256\times256$ resolution. 
We perform a user study to evaluate the performance of our EasyControl framework in the tasks of image-to-video and sketch-to-video generation. We employ VidRD as the foundational Text-to-Video (T2V) model, capable of generating 8-frame videos upon receiving conditions or text prompts at a resolution of $256\times256$.

\noindent \textbf{Comparison fairness}
% It is important to note that, due to the size of videos generated by SVD, VideoCrafter1, and I2Vgen-XL in the T2V task are different while the most evaluation condition data extracted from Midjourney~\cite{MidJourney} is equal in width and height, we set the width and height of the generation model equal to the smaller side. Furthermore, we keep the number of video frames generated by each model at the original default.
It's worth noting that, considering the varying sizes of videos generated by SVD, VideoCrafter1, and I2Vgen-XL in the Text-to-Video (T2V) task, while most evaluation condition data extracted from Midjourney~\cite{MidJourney} maintain equal width and height, we adjust the width and height of the generation model to match the smaller side. Additionally, we maintain the number of video frames generated by each model at the original default.

% \noindent \textbf{User study setting} The study involves 24 unique text-image pairs for I2V task and 20 unique text-sketch pairs for S2V task, and 30 participants are tasked with rating the output across these metrics. We use Likert scale~\cite{likert1932technique} surveys to analyze quantitative data, where users have the option to select from five rating levels, ranging from 1 (Extremely Dissatisfied) to 5 (Extremely Satisfied). More details of our user study are in the appendix. 
\noindent \textbf{User study setting} For the Image-to-Video (I2V) task, we curated 24 distinct text-image pairs, and for the Sketch-to-Video (S2V) task, we compiled 20 unique text-sketch pairs. We engaged 30 participants to evaluate the output based on predefined metrics. Utilizing Likert scale surveys~\cite{likert1932technique}, participants rated the generated content on a scale of 1 (Extremely Dissatisfied) to 5 (Extremely Satisfied). Additional details of our user study methodology are provided in the appendix.

\noindent \textbf{Evalution metrics}
\romannumeral1) \textbf{C-Match:} The alignment between the input conditions and the generated video.
\romannumeral2) \textbf{P-Match:} The alignment between the input prompt the the generated video.
\romannumeral3) \textbf{Consistency:} The temporal consistency of the generated video. 
\romannumeral4) \textbf{Quality:} Overall visual quality of the generated video.
% It focuses on the continuity and flow of visual elements, ensuring that the videos maintain a coherent visual structure.

\noindent \textbf{Results of Image-to-video task} The I2V user study results are presented in Tab~\ref{tab:t2v user study}, where our EasyControl achieves superior performance on each criterion against compared baselines. 
VideoComposer performs poorly in all metrics, particularly in the temporal consistency metric. This aligns with the findings in Sec~\ref{qual eval}, where we observe that videos generated by VideoComposer frequently suffer from flickering issues.
VideoCrafter1 and SVD exhibit greater temporal consistency than I2VGen-XL, which is in line with our conclusion from Tab.~\ref{tab:t2v metric}. We attribute I2VGen-XL's lower performance in C-Match and P-Match metrics compared to VideoCrafter1 and SVD to the absence of text guidance, which prevents the input of text prompts. Despite not being able to input text prompts, SVD generates videos with high fidelity and quality, resulting in high metrics for SVD.

\begin{table}[ht]
    % \small
    % \setlength{\tabcolsep}{2.8mm}
    \centering
    \caption{The result of the user study of image-to-video task conducted on 24 text-image pairs generated videos assessed by 20 human evaluators. 
    % There are 100 examples and 20 human raters.
    % Fidelity: similarity between image and video; Action: prompt action and video alignment; Content: prompt content and video alignment; Quality: visual perception of video; 
    }
    \vspace{-10pt}
    \begin{tabular}{lcccc}
       \toprule
       & C-Match & P-Match & Consistency & Quality \\
       \midrule
       VideoComposer & 2.97 & 2.93 & 2.43 & 2.80 \\
       I2VGen-XL & 3.28 & 3.31 & 3.29 & 3.09 \\
       VideoCrafter1 & 3.35 & 3.31 & 3.28 & 3.28 \\ 
       SVD & 3.38 & 3.37 & 3.37 & 3.28 \\
       \midrule
       Ours & \textbf{3.90} & \textbf{3.89} & \textbf{3.80} & \textbf{3.71} \\ 
       \bottomrule
    \end{tabular}
    
    \label{tab:t2v user study}
    \vspace{-8pt}
\end{table}

\noindent \textbf{Results of Sketch-to-video task}
The Tab~\ref{tab:s2vuser study} clearly shows a significant gap between VideoComposer and our method in terms of video quality, continuity, and match degree with conditions. This conclusion is further supported in Tab~\ref{tab:s2v metrics}, which demonstrates that the FVD and IS performance of VideoComposer on two datasets is significantly weaker than our model. We suspect that this may be due to the unstable control signal integrated with VideoComposer.

%##################################################################################################

\begin{table}[ht]
    \centering
    \caption{The result of the user study of sketch-to-video task conducted on 20 text-image pairs generated videos assessed by 20 human evaluators.
    }
    \vspace{-10pt}
    \begin{tabular}{lcccc}
       \toprule
       & C-Match & P-Match & Consistency & Quality \\
       \midrule
       VideoComposer & 2.35 & 2.93 & 1.69 & 1.44 \\
       Ours & \textbf{4.42} & \textbf{4.19} & \textbf{4.26} & \textbf{4.14} \\ 
       \bottomrule
    \end{tabular}
    
    \label{tab:s2vuser study}
    \vspace{-10pt}
\end{table}

%##################################################################################################

\section{Conclusion}
\label{sec:concl}

In this study, we introduce EasyControl, a versatile framework for controllable video generation. EasyControl accommodates diverse condition modalities such as image, canny edge, HED boundary, depth, sketch, and mask through training an additional condition adapter. This enhances control flexibility significantly. Leveraging our proposed condition propagation and injection schemes, our framework enables precise control using just a single condition map. Moreover, it facilitates training condition adapters with various pre-trained T2V models at a minimal cost. Extensive experimentation demonstrates the effectiveness and generalizability of our framework across diverse condition modalities and pre-trained T2V models. Results from quantitative analysis, qualitative assessments, and user studies collectively indicate that our approach surpasses existing state-of-the-art controllable video generation models in terms of both video quality and controllability. Notably, our method excels even in the image-to-video generation sub-task when compared to exclusive methods.

% In this work, we present DreamVideo, a model for synthesizing high-quality videos from images. Our DreamVideo has a great image retention capability and supports a combination of image and text inputs as controlling parameters.
% We propose an Image Retention block that combines control information and gradually integrates it into the primary U-Net. We Explore double-condition class-free guidance for different degrees of image retention. It's noteworthy that one limitation of our model is that the image retention ability of our DreamVideo relies on high-quality training data. Our DreamVideo is enabled to generate video that maintains superior image retention quality through training with high quality datasets. Finally, we demonstrate DreamVideo’s superiority over open-source image-video model qualitatively and quantitatively.
\clearpage

\bibliographystyle{IEEEtran}
\bibliography{main}

% \end{thebibliography}

% \newpage
 

% \end{IEEEbiography}



% \vfill

\end{document}